\documentclass[journal]{IEEEtran}
\usepackage[linesnumbered,ruled,vlined]{algorithm2e}

\usepackage{tablefootnote}
\usepackage{subcaption}
\usepackage{amsmath}
\usepackage{mathtools,amssymb,lipsum}

\usepackage{cuted}
\usepackage{dsfont}
\usepackage{amsmath}
\usepackage{msc}
\usepackage{tikz}
\usetikzlibrary{shapes,arrows}
\usepackage{tcolorbox}
\usepackage{ amssymb }
\usepackage{balance}
\usepackage{graphicx}
\usepackage[normalem]{ulem}
\usepackage{url,cite}
\usepackage{balance}
\usepackage{graphicx}
\usepackage{enumerate}
\usepackage{stmaryrd}
\usepackage{color}
\usepackage{multirow}
\usepackage{array}
\usepackage{float}
\usepackage[english]{babel}
\usepackage[utf8]{inputenc}
\usepackage[noend]{algpseudocode}
\newcolumntype{L}[1]{>{\raggedright\let\newline\\\arraybackslash\hspace{0pt}}m{#1}}
\newcolumntype{C}[1]{>{\centering\let\newline\\\arraybackslash\hspace{0pt}}m{#1}}
\newcolumntype{R}[1]{>{\raggedleft\let\newline\\\arraybackslash\hspace{0pt}}m{#1}}

\SetKwInput{KwInput}{Input}                
\SetKwInput{KwOutput}{Output}              

\usepackage{pifont}

\usepackage{amsmath}
\ifCLASSINFOpdf
\else
\fi
\begin{document}
%
\title{PLEX: Perturbation-free Local Explanations for LLM-Based Text Classification}
%
%
%

\author{Yogachandran~Rahulamathavan,
        Misbah~Farooq,
        and~Varuna~De~Silva

\thanks{
Y. Rahulamathavan, M. Farooq,   and V. Silva are with the Institute for Digital Technologies, Loughborough University London, London, U.K. (e-mails: \{y.rahulamathavan, M.Farooq, V.D.De-Silva\}@lboro.ac.uk).}
\thanks{
PLEX Sourcecode: https://github.com/rahulay1/PLEX.}


}

\maketitle

\begin{abstract}
Large Language Models (LLMs) excel in text classification, but their complexity hinders interpretability, making it difficult to understand the reasoning behind their predictions.  Explainable AI (XAI) methods like LIME and SHAP offer local explanations by identifying influential words, but they rely on computationally expensive perturbations. These methods typically generate thousands of perturbed sentences and perform inferences on each, incurring a substantial computational burden, especially with LLMs. To address this, we propose \underline{P}erturbation-free \underline{L}ocal \underline{Ex}planation (PLEX), a novel method that leverages the contextual embeddings extracted from the LLM and a ``Siamese network" style neural network trained to align with feature importance scores. This one-off training eliminates the need for subsequent perturbations, enabling efficient explanations for any new sentence. We demonstrate PLEX's effectiveness on four different classification tasks (sentiment, fake news, fake COVID-19 news and depression), showing more than 92\% agreement with LIME and SHAP.  Our evaluation using a ``stress test" reveals that PLEX accurately identifies influential words, leading to a similar decline in classification accuracy as observed with LIME and SHAP when these words are removed.  Notably, in some cases, PLEX demonstrates superior performance in capturing the impact of key features. PLEX dramatically accelerates explanation, reducing time and computational overhead by two and four orders of magnitude, respectively. This work offers a promising solution for explainable LLM-based text classification.
\end{abstract}


%
\IEEEpeerreviewmaketitle

\section{Introduction}
%
%
%
%
\IEEEPARstart{L}{arge} language models (LLMs) have significantly advanced text classification, achieving state-of-the-art results in tasks like emotion recognition, sentiment analysis, topic categorization, and spam detection \cite{ACMLLMSurvey}. Powered by transformer architectures with millions or billions of parameters, they effectively capture complex linguistic patterns.
However, the very complexity that enables their high performance also renders their internal workings opaque and difficult to interpret. This lack of transparency can hinder trust and acceptance, especially in critical applications like healthcare, finance, and legal domains \cite{ACMLLMSurvey}, where understanding the reasoning behind a model's decision is crucial for ensuring fairness, accountability, and reliability.  The need for interpretability arises not only to satisfy regulatory requirements or ethical considerations \cite{ACMLLMSurvey} but also to enable human users to gain insights from the model's predictions, debug potential errors, and ultimately make more informed decisions.

Explainable AI (XAI) methods have emerged to address the challenge of interpreting complex, black-box models by providing insights into their decision-making processes \cite{krishna2024,cantini2024,malhotra2024,hashmi2024, farooq2024,secondglance}. Local XAI methods, such as LIME (Local Interpretable Model-agnostic Explanations) \cite{LIME} and SHAP (SHapley Additive exPlanations) \cite{SHAP}, focus on explaining individual predictions. They achieve this by approximating the model's behavior within a localized region of the input space. For instance, consider a medical diagnosis model classifying a patient's description of their symptoms: \textit{``I've been experiencing persistent headaches, accompanied by nausea and blurred vision." }LIME or SHAP might reveal that the words ``headaches," ``nausea," and ``blurred vision" strongly contribute to a prediction of migraine. By identifying these influential words, LIME and SHAP provide transparency, facilitating trust in the AI system and offering medical professionals a deeper understanding of the model's diagnostic process. This allows clinicians to validate the model's prediction, identify potential biases, and make more informed decisions about patient care.

\begin{figure}\centering
\frame{\includegraphics[scale=0.35]{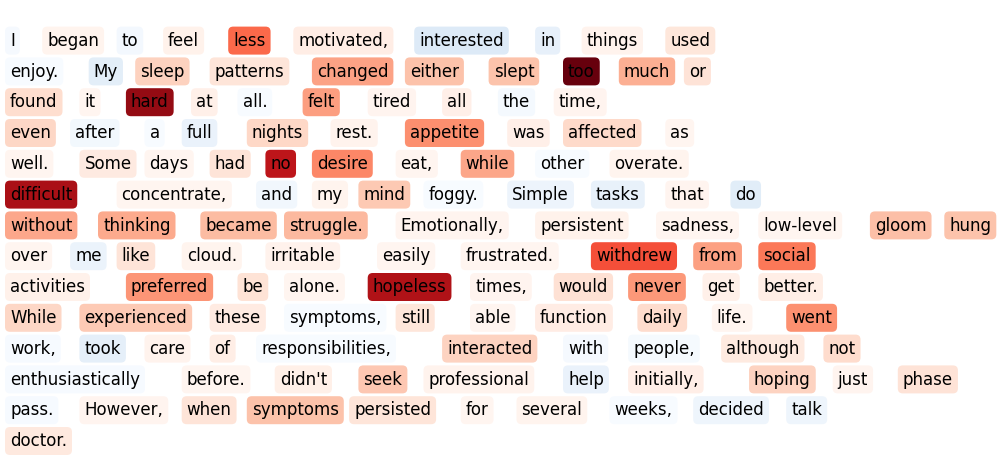}}
\caption{Visualization of word-level importance scores for an example sentence using the proposed PLEX approach. The sentence is classified as \textit{depression} by a fine-tuned BERT model. The PLEX framework effectively highlights both highly important and less important words, providing clear interpretability.}\label{Fig: ExamplHeatmap}
\end{figure}
LIME achieves this by generating perturbed versions of the original text, for instance, by removing or replacing words, and observing how these modifications affect the model's prediction.  Based on these observations, LIME constructs a simpler, interpretable model, often a linear model, that approximates the behavior of the complex model locally. This simpler model then highlights the words or phrases that are most influential in the original prediction, offering an explanation for the model's decision. SHAP, on the other hand, employs a game-theoretic approach to calculate the contribution of each word to the prediction. It assigns SHAP values to each word, quantifying their impact on the final prediction. These values can be positive or negative, indicating whether a word increases or decreases the likelihood of a particular class. 

While these methods have shown promise in explaining model predictions, they often rely on extensive perturbations, which can be computationally expensive and inefficient, particularly when applied to the complex architectures and vast parameter spaces of LLMs. This computational burden poses a significant challenge, especially in resource-constrained environments where generating thousands of perturbed samples and performing repeated inferences can be prohibitively time-consuming and expensive. The inherent complexity of LLMs further exacerbates this issue, as each inference requires substantial computational resources. This limitation motivates the exploration of alternative XAI methods that can provide efficient and interpretable explanations without relying on computationally intensive techniques like extensive perturbations.

To overcome this challenge, this paper introduces PLEX, a novel approach for \textbf{P}erturbation-free \textbf{L}ocal \textbf{Ex}planation of LLM-based text classification. PLEX leverages the rich contextual embeddings extracted from the LLM and employs a ``Siamese network" style neural network \cite{SiameseNetwork} trained to align these embeddings with feature importance scores.  In a one-shot learning fashion, this network learns a direct mapping between the embeddings and their corresponding importance, enabling efficient and interpretable explanations without the need for costly perturbations. Once trained, PLEX can efficiently generate explanations for any new sentence by directly leveraging the learned mapping, eliminating the computational burden associated with traditional XAI methods. 

To validate PLEX's effectiveness, we conducted rigorous experiments on four diverse tasks: sentiment prediction, fake news detection, COVID-19 fake news detection, and depression prediction. We employed fine-tuned BERT (Bidirectional Encoder Representations from Transformers) model \cite{BERT,T1,T2,T3,T4} to assess PLEX's generalizability across various language understanding tasks. For each task, PLEX was trained on specific datasets and underwent rigorous validation. To evaluate explanation performance, we subjected PLEX, LIME, and SHAP to a ``stress test," \cite{LIME} assessing the impact of removing top-contributing words on classification accuracy. Our results indicate that PLEX exhibits a high degree of agreement with both LIME and SHAP, achieving up to 92\% match across all four tasks. Additional tests further confirmed this agreement, highlighting PLEX's robustness and task-agnostic nature. Moreover, PLEX significantly outperforms LIME and SHAP in terms of computational efficiency, reducing overhead by four orders of magnitude. This makes PLEX a highly efficient and scalable solution for explaining LLM-based text classification models.

\subsection{Paper Organization}

The remainder of this paper is organized as follows. Section \ref{Section: LR} provides a comprehensive review of related work in the field of XAI for natural language processing (NLP), focusing on various techniques for interpreting text classification models. Section \ref{Section: Background} presents the necessary background on transformer encoders, the BERT architecture, and the process of fine-tuning LLMs for specific tasks. Section \ref{Section: Methodology} details our proposed PLEX method, including the proposed Siamese network architecture and the training process. Section \ref{Section: Experimental Setup} describes the experimental setup, including the datasets, evaluation metrics, and implementation details. Section \ref{Section: Results} presents the results of our experiments, comparing PLEX with existing XAI methods such as SHAP and LIME, and evaluating its performance. Finally, Section \ref{Section: Conclusions} concludes the paper by summarizing the key findings, discussing the implications of our work, and outlining potential future research directions.

\begin{table*}[h]
\centering
\caption{Comparison of XAI Methods for Text Classification}
\label{tab:xai-comparison}
\begin{tabular}{|c|c|c|c|c|c|c|c|}
\hline
\rotatebox{45}{\textbf{Method}} & \rotatebox{45}{\textbf{Explanation Type}} & \rotatebox{45}{\textbf{Model Agnostic}} & \rotatebox{45}{\textbf{Perturbation Based}} & \rotatebox{45}{\textbf{Surrogate Model}} & \rotatebox{45}{\textbf{Computational Cost}} & \rotatebox{45}{\textbf{Faithfulness}} & \rotatebox{45}{\textbf{Granularity}} \\ \hline
LIME \cite{LIME} & Local & Yes & Yes & Yes & High & High & Word \\ \hline
SHAP \cite{SHAP} & Local/Global & Yes & Yes & No & High & High & Word \\ \hline
AttentionViz \cite{yeh2023} & Global & No & No & No & Moderate & Moderate & Sentence \\ \hline
IGs \cite{sundararajan2017} & Local & No & No & No & Low & Low (for LLMs) & Word \\ \hline
TokenSHAP \cite{horovicz2024} & Local & Yes & Yes & No & High & High & Word \\ \hline
LRP \cite{bach2015} & Local & No & No & No & Moderate & Low (for LLMs) & Word \\ \hline
TEXTFOOLER \cite{jin2020} & Local & Yes & Yes & No & High & Moderate & Word \\ \hline 
UEKD \cite{liu2024} & Local & No & No & No & Moderate & High & Word \\ \hline 
TalkToModel \cite{slack2023} & Local & Yes & Yes & No & High & Moderate & Sentence \\ \hline 
dEFEND \cite{shu2019} & Local & No & No & No & Moderate & High & Sentence \\ \hline 
MTBERT-Attention \cite{sebbaq2023} & Local & No & No & No & Moderate & Moderate & Word \\ \hline 
MARTA \cite{arous2021} & Local & No & Yes & No & High & Moderate & Word \\ \hline 
\textbf{PLEX (Ours)} & Local & No & No & No & Low & High & Word \\ \hline
\end{tabular}
\end{table*}

\section{Literature review }\label{Section: LR}
To build trustworthy AI models, it is essential to balance the need for accurate predictions with the ability to explain those predictions. XAI addresses this challenge by providing insights into how these models works, transforming them from opaque black boxes into explainable models. Local and global explanations are two primary techniques for explaining fine-tuned LLMs \cite{ACMLLMSurvey}. Local explanations focus on understanding the model's decision-making process for individual inputs \cite{LIME}, while global explanations aim to provide a broader understanding of the model's overall behaviour \cite{LIME,SHAP}.  This work focuses on local explanations. Local explanations can be categorized into three primary approaches: feature attribution-based explanation, example-based explanation, and attention-based explanation. The relevant works are compared in Table \ref{tab:xai-comparison}.

\subsection{Feature-Attribution Based Explanation}

This approach assigns importance scores to input words to explain model predictions. Techniques like SHAP \cite{SHAP} and LIME \cite{LIME} are widely used across tasks such as sentiment analysis, misogyny detection, and code summarisation \cite{wu2021, attanasio2022, li2024} and rely on perturbation analysis. For example, \cite{wu2021} compared attribution techniques for BERT, finding that while gradient-based methods and Layer-wise Relevance Propagation (LRP) offered consistent results, attention-based approaches were less reliable.

Gradient-based methods, however, can struggle in deep architectures like LLMs due to issues such as vanishing gradients. \cite{attanasio2022} showed their limitations in misogyny detection tasks, where SHAP and Sampling and Occlusion (SOC) produced more faithful explanations. Recent methods like Attributive Masking Learning (AML) \cite{barkan2024} and TokenSHAP \cite{horovicz2024} enhance attribution quality using masking strategies or Shapley values.

Despite these advances, many attribution methods remain computationally expensive. In contrast, our PLEX method directly maps LLM embeddings to importance scores without perturbations, enabling faster, scalable, and more efficient explanations.

\subsection{Example-Based Explanations}

Example-based explanations clarify model behavior through illustrative instances such as adversarial or counterfactual examples. TEXTFOOLER \cite{jin2020} generates adversarial inputs via synonym substitutions, but this can distort meaning and incur high computational cost. Similarly, generating quality counterfactuals is challenging.

Recent works such as UEKD \cite{liu2024} and LLAMOS \cite{lin2024} improve robustness, while tools like TalkToModel \cite{slack2023} support interactive model exploration. However, ensuring validity and fidelity of generated examples remains difficult.

\subsection{Attention-Based Explanations}

Attention-based methods utilize transformer attention weights to infer token relevance. Models like AttentionViz \cite{yeh2023} and keyword-based classifiers \cite{du2018} demonstrate their utility. However, studies \cite{liu2021, shu2019} caution that attention weights may not reliably reflect true feature importance and can be misleading \cite{bastings2020}.

Our PLEX framework avoids these pitfalls by not relying on attention weights. Instead, it learns a mapping from contextual embeddings to importance scores using a dedicated “Seismic network,” offering a more faithful and interpretable alternative to traditional attention-based explanations.

\section{Background Information}\label{Section: Background}
\subsection{Transformer Encoders}
Transformer encoders are a type of neural network architecture that have revolutionized NLP due to their ability to capture long-range dependencies and contextual information in sequential data \cite{attentionisallyouneed}. Unlike traditional recurrent neural networks (RNNs), which process input tokens sequentially, transformers utilize a mechanism called self-attention to weigh the importance of different tokens in the input sequence when encoding a specific token. This allows the model to consider the relationships between all tokens in the sequence simultaneously, leading to a more comprehensive understanding of the context.

The core component of a transformer encoder is the attention mechanism, which calculates attention scores between each pair of tokens in the input sequence. These scores represent the relevance of one token to another when encoding their respective representations. The attention scores are then used to weight the values of the tokens, effectively allowing the model to focus on the most relevant parts of the input when encoding a specific token. This self-attention mechanism enables transformers to capture long-range dependencies and contextual information more effectively than RNNs, leading to significant performance improvements in various NLP tasks, including text classification.
\subsection{BERT Architecture}
BERT is a transformer-based language model that builds upon the encoder architecture \cite{BERT}. It consists of multiple stacked transformer encoder layers, each incorporating self-attention mechanisms to capture contextual relationships between words in a sentence. BERT's key innovation lies in its bidirectional training, where the model learns to predict masked words in a sentence based on both the preceding and following context. This bidirectional approach allows BERT to generate contextualized word embeddings that capture richer semantic and syntactic information compared to traditional word embeddings.

The input to BERT is a sequence of tokens, typically representing words or subwords. Each token is embedded into a vector representation, which is then processed by the transformer encoder layers. The output of BERT is a sequence of contextualized embeddings, one for each input token. These embeddings can be used for various downstream NLP tasks, including text classification, question answering, and named entity recognition. BERT's ability to generate highly contextualized representations has led to significant performance improvements across a wide range of NLP applications, making it a foundational architecture for many modern language models. The BERT base model comprises 12 transformer encoder layers, each with a hidden size of 768, 12 self-attention heads, and a feed-forward size of 3072. This architecture results in a model with approximately 110 million parameters.  BERT's training was conducted on a massive text corpus combining the BooksCorpus (800 million words) and English Wikipedia (2,500 million words).  

\subsection{Fine-tuning BERT for Sentiment Classification}
While BERT's pre-training on a massive text corpus equips it with general language understanding capabilities, fine-tuning is crucial for adapting it to specific downstream tasks like sentiment classification. Fine-tuning involves further training the pre-trained BERT model on a labelled dataset specific to the target task. In the case of sentiment classification, this dataset would consist of text samples paired with their corresponding sentiment labels (e.g., positive, negative, neutral).

During fine-tuning, the entire BERT architecture is typically trained, allowing the model to adjust its parameters and learn task-specific representations. A key modification for classification tasks is the addition of a classification layer on top of the final transformer encoder layer. This classification layer takes the output embedding of the [CLS] token, which represents the aggregated sentence representation, and maps it to a probability distribution over the classes of the tasks. The model is then trained using a cross-entropy loss function to optimize its performance.

Fine-tuning BERT for classification tasks offers several advantages. It allows the model to leverage its pre-trained knowledge while adapting to the nuances of the specific task. This process often leads to significant improvements in classification accuracy compared to training a model from scratch. Additionally, fine-tuning enables the model to capture the specific vocabulary and linguistic patterns associated with the classification task in the target domain, resulting in more robust and accurate performance.

\subsection{LIME and SHAP based Explanation}

\begin{figure}[!t]\centering
\frame{\includegraphics[trim=0cm 3cm 2cm 4cm, clip, scale=0.27]{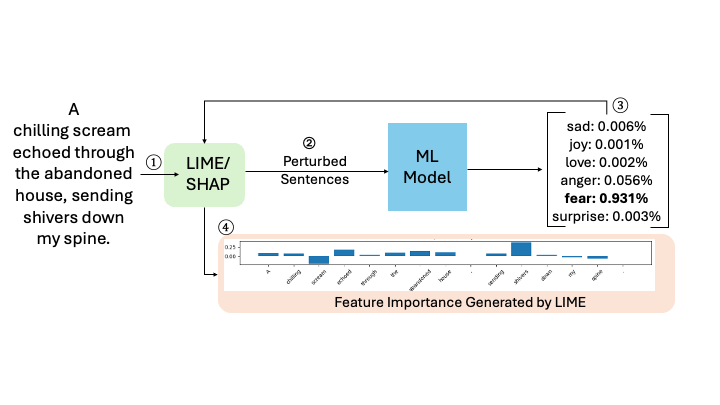}}
\caption{Block diagram showing the process of getting the word importance for a given sentence using LIME and SHAP.}\label{Fig: BlockDiaExample}
\end{figure}

Figure \ref{Fig: BlockDiaExample} illustrates a typical pipeline for emotion classification followed by explanation using LIME or SHAP. The process begins with a sentence being input to the model, which produces a probability distribution over a set of emotions (e.g., joy, sadness, anger). The emotion with the highest probability is then identified as the predicted emotion. To explain this prediction, LIME or SHAP generates a large number of perturbed sentences by systematically modifying the original input. These perturbed sentences are then fed back into the model, and the resulting changes in prediction accuracy are analyzed to determine the importance of each word in the original sentence. This process allows for a localized understanding of the model's decision-making process, highlighting the words that most significantly contribute to the predicted emotion. 

\section{Methodology}\label{Section: Methodology}

This section details the methodology employed in developing and evaluating PLEX, our novel approach for perturbation-free local explanations in LLM-based text classification. We first describe the process of constructing the training dataset, which involves leveraging LIME and SHAP to obtain feature importance scores for a diverse set of sentences. Subsequently, we outline the architecture and training procedure of the Siamese network, which forms the core of the PLEX method. In Section \ref{Section: Experimental Setup}, we discuss the evaluation metrics and experimental setup used to assess the performance and efficiency of PLEX in comparison to traditional XAI methods.

\subsection{Intuition behind the work}

\begin{figure}[!h]\centering
\frame{\includegraphics[trim=0cm 0cm 0cm 0cm, clip, scale=0.32]{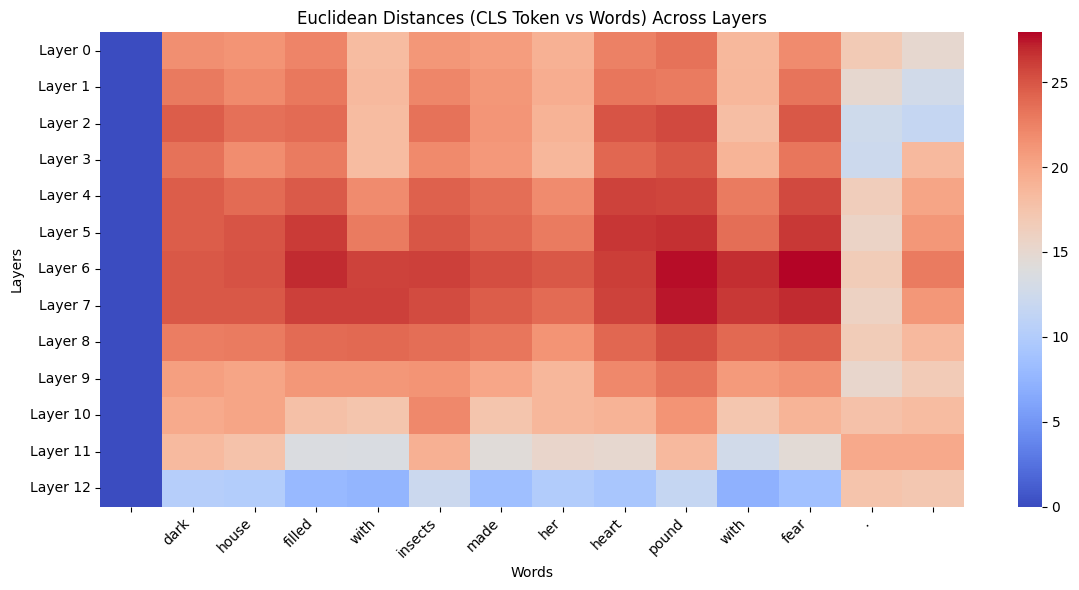}}
\caption{Heatmap showing the Euclidean distances between CLS token embeddings and word embeddings for all layers of BERT.}\label{Fig: DirectCorr}
\end{figure}

Our approach is motivated by the observation that the CLS token embedding, a representation of the entire sentence generated by BERT, plays a pivotal role in classification. We hypothesise that individual word embeddings within a sentence should correlate with the CLS token embedding, reflecting their contribution to the overall sentiment. This correlation was explored by measuring Euclidean distances between the CLS token embedding and individual word embeddings across all 12 layers of the fine-tuned BERT model, as visualized in Figure \ref{Fig: DirectCorr}.  

\begin{figure}[!h]\centering
\frame{\includegraphics[trim=3cm 3cm 2cm 2cm, clip, scale=0.3]{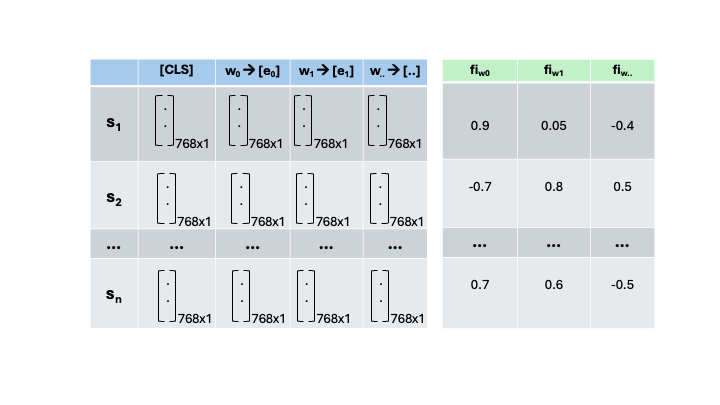}}
\caption{Creation of Dataset for training the Siamese network. Each row represent a sentence. The blue columns contain the embeddings for CLS token and other words in the sentence. The Green columns contain the word importance score for each word in the sentence.}\label{Fig: Dataset}
\end{figure}

As the words progress through the layers, the similarity between their embeddings and the CLS token embedding generally increases, particularly in the final layers. For instance, in the example sentence \textit{``dark house filled with insects made her heart pound with fear,"} we expect that the words ``dark," ``insects," and ``fear" exhibit a stronger correlation with the CLS token embedding, aligning with their significant contribution to the expressed emotion.

\begin{figure*}[!ht]\centering
\frame{\includegraphics[trim=0cm 0cm 0cm 0cm, clip, scale=0.17]{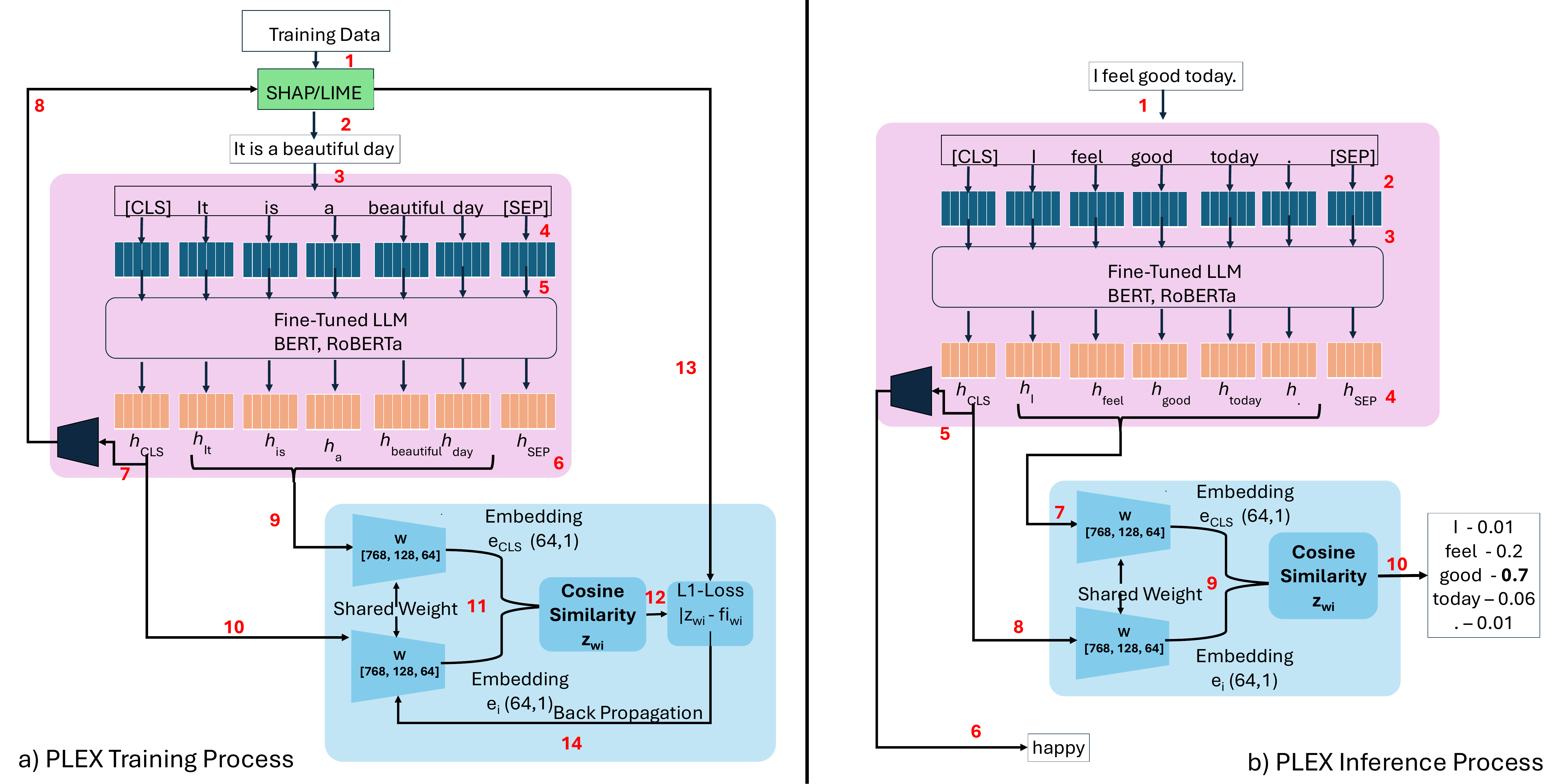}}
\caption{PLEX Architecture: Training and Inference Processes. The pseudo codes are provided in Algorithms 1 and 2.}
\label{Fig: BlockDia}
\end{figure*}

To further investigate this hypothesis, we employed LIME and SHAP to obtain feature importance scores for each word. While we observed a positive correlation between these scores and the similarity between word embeddings and the CLS token embedding, this correlation alone proved insufficient to fully explain the sentiment of the entire text, as depicted in Figure \ref{Fig: DirectCorr}. This finding motivated the development of our proposed Seismic network to refine and enhance this correlation.

\subsection{Dataset Construction}

To strengthen the correlation between CLS and word tokens, we constructed a novel dataset comprising CLS token embeddings, word embeddings, and corresponding word importance scores, as illustrated in Figure \ref{Fig: Dataset}.  This dataset leverages the semantic representations captured by the embeddings and their respective contributions to the overall sentiment, providing the necessary information to train the Siamese network.

To facilitate a comprehensive analysis, we constructed two distinct datasets for each classification task. These datasets differ in the method employed to generate feature importance scores: one utilizes LIME, while the other employs SHAP. This dual approach allows us to investigate the influence of different feature attribution methods on the performance and explainability of our proposed Siamese network. With approximately 400 sentences per class and an average sentence length of 30 words, each dataset comprises around 12000 word embeddings paired with their corresponding importance scores.  Crucially, this method of labeling word embeddings with their importance, derived from established XAI techniques, is significantly faster and more efficient than manual labeling by human annotators. This data captures the intricate relationship between word embeddings and their contribution to the predicted sentiment, providing the necessary foundation for training the Siamese network.

\subsection{Siamese Network Architecture and Training}

To enhance the correlation between the CLS token and word token embeddings, we employ a Siamese network architecture \cite{SiameseNetwork}.  Inspired by their success in image recognition tasks, we adapt Siamese networks to operate in the embedding space, aiming to increase the similarity between the CLS token embedding ($\mathbf{h}_{cls}$) and word embeddings ($\mathbf{h}_{w}$) for words that contribute positively to the classification decision, while reducing similarity for those that contribute negatively.

Our Siamese network consists of two identical subnetworks, each comprising two fully connected layers with ReLU activation and dropout for regularization. As depicted blue boxes in Figure \ref{Fig: BlockDia}, the network takes $\mathbf{h}_{cls}$ and $\mathbf{h}_{w}$ as input and produces outputs $\mathbf{e}_{cls}$ and $\mathbf{e}_{w}$, respectively, where:
\begin{equation}\nonumber
\mathbf{e}_{cls} = \mathbf{W} (\mathbf{h}_{cls}), and~~
\mathbf{e}_w = \mathbf{W}(\mathbf{h}_{w}).
\end{equation}
Here, $\mathbf{W}$ represents the transformation function of the neural network.  We then compute the cosine similarity between $\mathbf{e}_{cls}$ and $\mathbf{e}_{w}$:

\begin{equation*}
\text{sim}(\mathbf{e}_{cls}, \mathbf{e}_{w}) = \frac{\mathbf{e}_{cls} \cdot \mathbf{e}_{w}}{\|\mathbf{e}_{cls}\| \|\mathbf{e}_{w}\|}.
\end{equation*}
This similarity score, $\text{sim}(\mathbf{z}_{cls}, \mathbf{z}_{w})$, is compared to the corresponding word's feature importance obtained from LIME/SHAP ($fI_w$) from the constructed dataset (Figure \ref{Fig: Dataset}).  Cosine similarity effectively captures the semantic alignment between the CLS token embedding and individual word embeddings, providing a normalized score between -1 and 1 that aligns with the range of feature importance scores.

To train the network, we utilize the L1-loss (mean absolute error) to measure the difference between the predicted similarity and the target feature importance:

\begin{equation*}
\mathcal{L} = \alpha.|fI_w|.|\text{sim}(\mathbf{z}_{cls}, \mathbf{z}_{w}) - fI_w|,
\end{equation*}
where $\alpha$ is a scaling factor and $|fI_w|$ prioritizes the most and least important words during training. This loss is backpropagated through the network to update its weights, enabling it to learn a mapping between embeddings and their importance scores.  The absence of an activation function in the final layer allows the network to predict both positive and negative similarity values, consistent with the nature of LIME/SHAP feature importance. Refer to Figure \ref{Fig: BlockDia}, Algorithms 1 and 2 for more details.

\begin{algorithm}[htbp]
\caption{PLEX Training Process}
\KwIn{Dataset $\mathcal{D}$ of sentences, Fine-tuned LLM, SHAP/LIME, Siamese Network}
\KwOut{Trained Siamese Network}

    Randomly select a sentence $S$ from $\mathcal{D}$\ and send $S$ to SHAP/LIME for perturbation\;
    Obtain perturbed sentences $S_{\text{perturbed}}$\;
    Tokenize $S_{\text{perturbed}}$ into tokens $T$\;
    Generate token embeddings $E_T$ for $T$\;
    Pass $E_T$ through the fine-tuned LLM\;
    Obtain contextualized embeddings $E_C$\;
    Extract CLS token embedding $E_{\text{CLS}}$ from $E_C$ and
    send $E_{\text{CLS}}$ to the classifier to predict probability distribution $P_{\text{class}}$\;
    Provide $P_{\text{class}}$ to SHAP/LIME for word importance estimation\;
    Send $E_C$ (contextualized word embeddings) to the Siamese network\;
    Send $E_{\text{CLS}}$ (contextualized CLS embedding) to the Siamese network\;
    Compute outputs $O_S$ of the Siamese network\;
    Calculate cosine similarity $\text{sim}(O_S)$\;
    Retrieve word importance scores $fi_{\text{wi}}$ from SHAP/LIME\;
    Compute the L1-loss $L$ and backpropagate to update the Siamese network parameters\;
\end{algorithm}

\begin{algorithm}[htbp]
\caption{PLEX Inference Process}
\KwIn{Sentence, Fine-tuned LLM,  Trained Siamese Network}
\KwOut{Word Importance Scores}
    Tokenize the sentence into tokens $T$\;
    Generate token embeddings $E_T$ for $T$\;
    Pass $E_T$ through the fine-tuned LLM \;
    Obtain contextualized embeddings $E_C$\;
    Extract CLS token embedding $E_{\text{CLS}}$ from $E_C$\;
    Send $E_{\text{CLS}}$ to the classifier to predict probability distribution $P_{\text{class}}$\;
    Send $E_C$ (contextualized word embeddings) to the Siamese network\;
    Send $E_{\text{CLS}}$ (contextualized CLS embedding) to the Siamese network\;
    Compute outputs $O_S$ of the Siamese network\;
    Calculate cosine similarity $\text{sim}(O_S)$ and output the word importance score\;
\end{algorithm}


\section{Experimental Setup}\label{Section: Experimental Setup}
To validate the proposed PLEX algorithm, we consider four text classification tasks. To enable this experiment we use four fine-tuned BERT models as described below:
\begin{itemize}
    \item Emotion Classification: This task involves classifying text into six basic emotions: sadness, joy, love, anger, fear, and surprise. We utilize a BERT model fine-tuned on a Twitter dataset for this task \cite{T1}.
    \item Fake News Detection: This task addresses the growing problem of misinformation by classifying news articles as real or fake. We employ a RoBERTa model fine-tuned on the FakeNewsNet dataset \cite{T2}.
    \item COVID-19 Fake News Detection: This task focuses specifically on identifying fake news related to the COVID-19 pandemic. We use another RoBERTa model fine-tuned on a COVID-19 fake news dataset \cite{T3}.
    \item Depression Classification: This task deals with mental health by classifying text as indicative of depression or not. We utilize a BERT model fine-tuned on a mental health dataset \cite{T4}.
\end{itemize}

Table \ref{tab:bert-models} presents a summary of the fine-tuned BERT models employed for the four tasks considered in this paper. Each model is characterized by its base architecture, tokenizer, number of parameters, floating-point operations (FLOPs) per word, embedding size, and number of layers. The models range from smaller, general-purpose BERT models fine-tuned for tasks such as emotion classification and depression detection to larger, task-specific models like those used for COVID-19 fake news detection.  In the following subsection, we describe the datasets we used for all four tasks in detail. The fine-tuned BERT models exhibit strong performance across all tasks, achieving accuracy above 90\% in all four cases. 

\begin{table*}[!h]
\centering
\caption{Fine-tuned BERT Models Used in the Study.}
\label{tab:bert-models}
\begin{tabular}{|c|c|c|c|c|c|c|c|}
\hline
\textbf{\rotatebox{45}{Fine-tuning Task}} & 
\textbf{\rotatebox{45}{Base \newline Model}} & 
\textbf{\rotatebox{45}{Tokenizer}} & 
\textbf{\rotatebox{45}{Parameters \newline (M)}} & 
\textbf{\rotatebox{45}{FLOPs per \newline word (G)}} & 
\textbf{\rotatebox{45}{Embedding \newline Size}} & 
\textbf{\rotatebox{45}{Layers}} & 
\textbf{\rotatebox{45}{Accuracy}} \\ \hline
T1. Emotion Classification \cite{T1} & BERT & BertTokenizerFast & 110 & 0.26 & 768 & 12 & $93.4\%$ \\ \hline
T2. Fake News Detection \cite{T2} & RoBERTa & RobertaTokenizerFast & 125 & 0.34 & 768 & 12 & $96.2\%$ \\ \hline
T3. COVID-19 Fake News Detection \cite{T3} & RoBERTa & RobertaTokenizerFast & 355 & 1.21 & 1024 & 24 & $94.5\%$ \\ \hline
T4. Depression Classification \cite{T4} & BERT & BertTokenizerFast & 110 & 0.26 & 768 & 12 & $92.8\%$ \\ \hline
\end{tabular}
\end{table*}

\subsection{Datasets}
This section provides a detailed overview of the datasets used in the study, along with the corresponding performance metrics of the fine-tuned BERT models. For each task, the number of classes, the dataset source, the training and test set sizes, the average number of words per sample, the achieved accuracy, the number of training epochs, and the final loss are presented in  Table \ref{tab:dataset-stats}.

The datasets vary in size, complexity, and domain. The emotion classification task utilizes a Twitter dataset with six classes (sad, joy, love, anger, fear or surprise) \cite{datasetTwitter}, Fake New Detection task utilises FakeNewsNet dataset with two classes (fake or real)\cite{datasetFakeNews}, Covid-19 Fake News detection task utilises Covid-19 Fake News Dataset with two classes (fake or real) \cite{datasetCovidFakeNews} and depression classification tasks utilises Reddit Depression Dataset with two classes (depression or normal) \cite{datasetDepression}.

\begin{table*}[!h]
\centering
\caption{Dataset Statistics and Model Performance}
\label{tab:dataset-stats}
\begin{tabular}{|c|c|c|c|c|c|c|c|c|}
\hline
\textbf{Task} & \textbf{Classes} & \textbf{Dataset} & \textbf{Train} & \textbf{Pairs} &\textbf{Test} & \textbf{Avg. Words} &  \textbf{Epochs} & \textbf{Loss} \\ \hline
Emotion Classification & 6 & Huggingface Twitter \cite{datasetTwitter} & 2400 & 50000 & 1200 & 22 &  400 & 0.04 \\ \hline
Fake News Detection & 2 & FakeNewsNet \cite{datasetFakeNews} & 800 & 30000 & 400 & 35  & 450 & 0.05 \\ \hline
COVID-19 Fake News Detection & 2 & COVID-19 Fake News \cite{datasetCovidFakeNews} &  800 & 25000& 400 & 30 & 500 & 0.05 \\ \hline
Depression Classification & 2 & Reddit Depression Dataset \cite{datasetDepression} &  800 & 32000& 400 & 40 & 450 & 0.04 \\ \hline
\end{tabular}
\end{table*}

\subsection{Novel Dastset Creation}
To train the Siamese network, we construct a custom dataset as detailed in Section \ref{Section: Methodology}. This dataset comprises 400 sentences per class for each task, resulting in 2400 sentences for emotion detection. Given an average sentence length of approximately 30 words, each sentence yields 30 CLS-word pairs for training. Consequently, the ``Pairs" column in Table \ref{tab:dataset-stats} reflects the effective size of the training data.

To train the Siamese network, we construct a custom dataset as detailed in Section \ref{Section: Methodology}. This dataset comprises 400 sentences per class for each task, resulting in 2400 sentences for emotion detection. Given an average sentence length of approximately 30 words, each sentence yields 30 CLS-word pairs for training. Consequently, the ``Train" column in Table \ref{tab:dataset-stats} reflects the effective size of the training data.

For each word in a sentence, we require an importance score between $-1$ and $1$, indicative of its contribution to the sentence classification.  For example, in the sentence \textit{``This movie is absolutely phenomenal!"}, the word \textit{``phenomenal"} might have an importance score of $0.8$, while the word \textit{``is"} might have a score of $0.1$, as illustrated in Figure \ref{Fig: Dataset}. Manually assigning these scores for all the sentences would be time-consuming. Therefore, we use LIME and SHAP to obtain these scores.

\subsubsection{LIME-based dataset}
To generate the LIME-based dataset, we leverage the LIME framework \cite{LIME} to explain the predictions from the fine-tuned BERT model for each sentence. This process, depicted in Figure \ref{Fig: BlockDia}, involves several key steps. Initially, the original sentence is input to the LIME explainer. LIME then generates a set of perturbed instances by randomly removing words from the original sentence. These perturbed instances, along with the original sentence, are then fed to the BERT model to obtain their respective predicted probabilities for each class. Subsequently, LIME employs these predictions to train a local surrogate model, typically a linear model, which approximates the behaviour of the BERT model in the vicinity of the original sentence. Finally, LIME extracts the coefficients of this surrogate model, representing the feature importance scores for each word in the original sentence. These importance scores, ranging from $-1$ to $1$, quantify the contribution of each word to the predicted sentiment.

\subsubsection{SHAP-based dataset}

The SHAP-based dataset is generated using the SHAP framework, which employs a game-theoretic approach to explain predictions. As illustrated in Figure \ref{Fig: BlockDia}, the process involves similar steps to LIME. The original sentence is sent to the SHAP explainer, which then generates perturbed versions by systematically removing words. These perturbed sentences are passed through the BERT model to obtain their prediction probabilities. SHAP then calculates the contribution of each word to the prediction by considering all possible combinations of words and their impact on the output. This results in SHAP values for each word, representing their contribution to the predicted sentiment. These SHAP values, also ranging from $-1$ to $1$, are used as the feature importance scores in the SHAP-based dataset.

\subsection{Training the Siamese Network}
For each of the four text classification tasks, we trained two distinct Siamese network models, resulting in a total of eight trained models. We denote the model trained using the LIME-based dataset as PLEX-LIME and the model trained using the SHAP-based dataset as PLEX-SHAP. 

To construct the training dataset for our Siamese network, we adopted a strategy that prioritizes both randomness and representation.  Initially, we compiled all possible CLS-word embedding pairs along with their corresponding LIME or SHAP importance scores from the entire training set.   Next, we randomized this collection to ensure that the network is not biased by the order of sentences or words in the original dataset. 

Finally, we divided this shuffled collection into batches of $32$ pairs each. This batching strategy serves two primary purposes: first, it facilitates efficient training by processing multiple samples simultaneously, and second, it introduces randomness within each batch, exposing the network to a varied set of word-CLS relationships and promoting better generalization. Table \ref{tab:dataset-stats} and Table \ref{tab:comp-cost} shows the number of epochs, loss and feature extraction and training times. Feature extraction is much faster with LIME than SHAP. However, the training times are similar for both the datsets.

\begin{table*}[!h]
\centering
\caption{Computational Cost For Training the PLEX Models}
\label{tab:comp-cost}
\begin{tabular}{|c|c|c|c|c|}
\hline
\textbf{Task} & \multicolumn{2}{c|}{\textbf{Feature Extraction (hours)}} & \multicolumn{2}{c|}{\textbf{Training (hours)}} \\ \hline
 & \textbf{LIME} & \textbf{SHAP} & \textbf{PLEX-LIME} & \textbf{PLEX-SHAP} \\ \hline
Emotion Classification & $\approx$1 & $\approx3$ & $\approx$6 &$\approx$6 \\ \hline
Fake News Detection & $\approx$1 & $\approx$3 & $\approx$6 & $\approx$6 \\ \hline
COVID-19 Fake News Detection & $\approx$1 & $\approx$3 & $\approx$6 & $\approx$6 \\ \hline
Depression Classification & $\approx$1 & $\approx$3 & $\approx$6 & $\approx$6 \\ \hline
\end{tabular}
\end{table*}

Our Siamese network architecture consists of two identical subnetworks, each designed to process an input embedding. The input layer size varies depending on the task: 768 dimensions for tasks 1, 2, and 4, and 1024 dimensions for task 3. This difference accommodates the varying dimensionality of the pre-trained language models used for each task. Following the input layer, each subnetwork comprises two fully connected layers with 128 and 64 units, respectively.  ReLU activation functions are applied to the first fully connected layer to introduce non-linearity. Additionally, a dropout layer with a rate of 50\% is employed after the first fully connected layer to mitigate overfitting and improve generalization.

\begin{figure*}[!h]\centering
\frame{\includegraphics[scale=0.45]{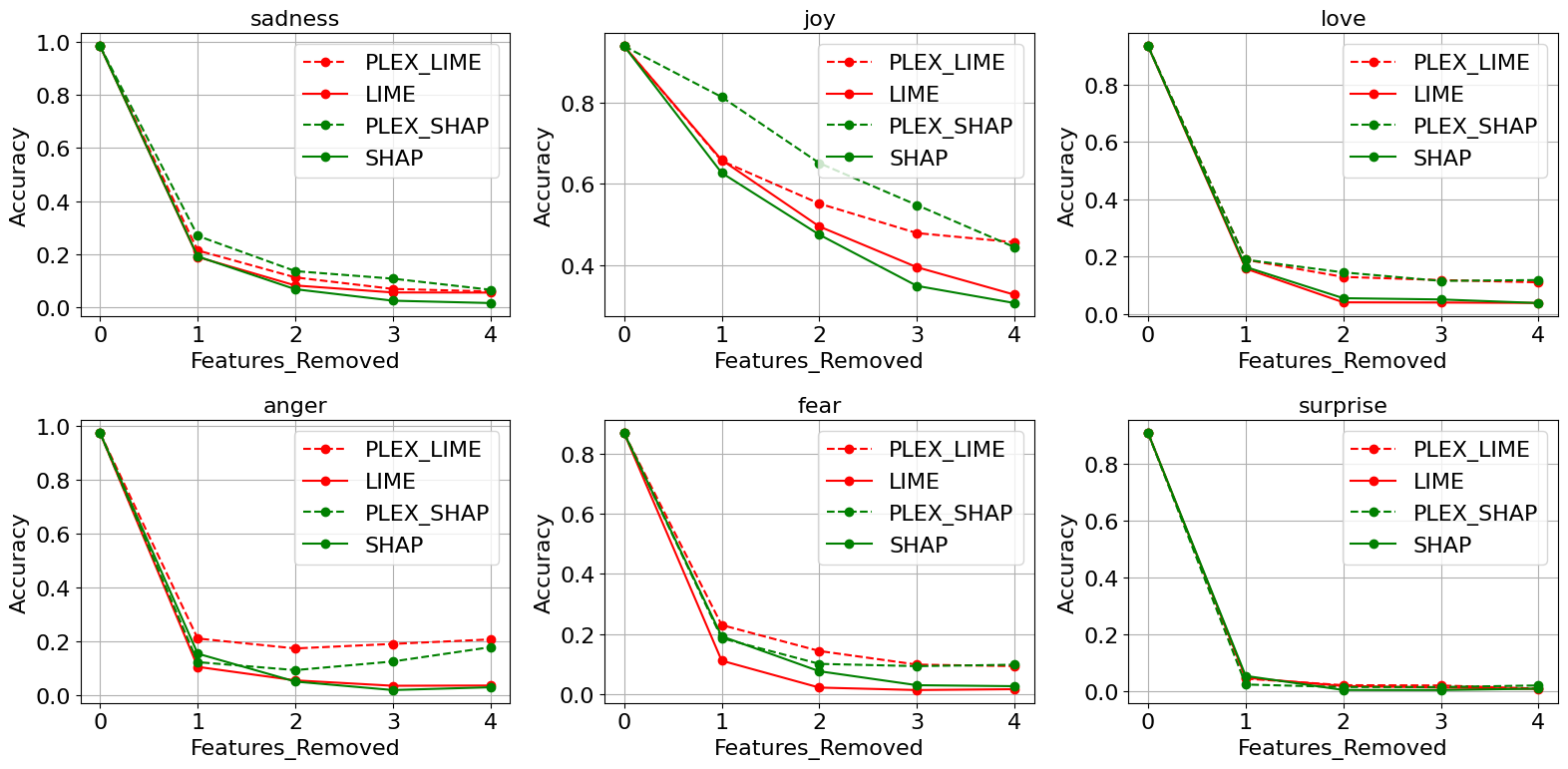}}
\caption{Stress test to compare the accuracy reduction when removing the words based on their importance.}
\label{Fig: allemotionStressTest}
\end{figure*}

\begin{figure*}[!h]\centering
\frame{\includegraphics[trim=0cm 18cm 1cm 10cm, clip, scale=0.18]{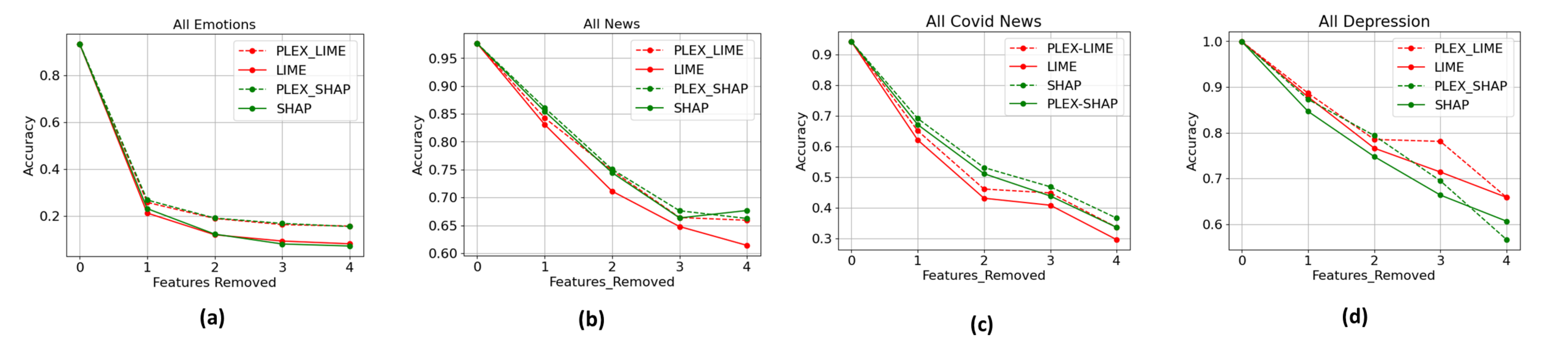}}
\caption{Stress test results comparing the accuracy reduction when removing words based on their importance across four tasks: (a) Emotion Task, (b) Fake News Task, (c) Covid Fake News Task, and (d) Depression Task. }\label{Fig: StressTestalluseecases}
\end{figure*}

\section{Experimental Results}\label{Section: Results}
\subsection{Stress Test}
Within the context of XAI, a ``stress test" is a technique used to evaluate the faithfulness and stability of explanations generated by an XAI method. It involves systematically perturbing the input features (e.g., words in text classification tasks) and observing the impact on the model's prediction. By analyzing how the prediction changes in response to these perturbations, one can assess whether the explanation accurately captures the features that are truly influential in the model's decision-making process. For example, if an XAI method highlights certain words as being important for a sentiment classification task, a stress test might involve removing those words and observing whether the model's predicted sentiment changes accordingly. A significant change in the prediction would suggest that the explanation is faithful, as it correctly identified the influential features. 

In this work, we employ a stress test to evaluate the faithfulness of explanations generated by the proposed PLEX-SHAP, and PLEX-LIME models aganist the state-of-the-art SHAP and LIME explanations. For each method, we systematically remove the top \textit{k} most important words identified by the respective explainer from the input text and observe the resulting change in the model's predicted probability for the original class. This process is repeated for \textit{k} ranging from 1 to 4, allowing us to assess the impact of removing increasingly larger sets of influential words. By comparing the accuracy decline patterns across the four methods, we can evaluate their relative faithfulness in identifying the truly important features driving the model's prediction.

Figure \ref{Fig: allemotionStressTest} shows the stress test results for emotion detection usecase for all six emotions. As expected, removing important words leads to a decline in classification accuracy across all methods and emotions.  When removing words based on LIME or SHAP feature importance scores, the observed decline defines a lower bound, as LIME and SHAP are specifically designed to identify the most influential words for the prediction.

As expected, the accuracies of our proposed PLEX-LIME and PLEX-SHAP models decline in a similar fashion as SHAP and LIME models. This observation reinforces the notion that our method successfully identifies and prioritizes features crucial for sentiment prediction, aligning with LIME and SHAP's explanations. We observed a similar trend across all four tasks as shown in Figure \ref{Fig: StressTestalluseecases} where the loss of accuracy of the proposed method closely follows the loss of accuracy of SHAP and LIME based approaches. At some occasions, the proposed approach outperforms SHAP and LIME i.e., anger and surprise classes in emotion detection tasks.

\subsection{Evaluation of explanation agreement}
\begin{figure}[!h]\centering
\frame{\includegraphics[scale=0.26]{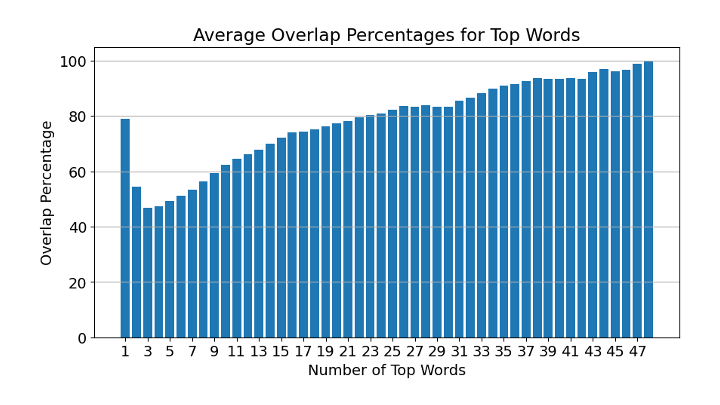}}
\caption{Average overlap percentages between the top k important words identified by the model and LIME, calculated across all sentences in the dataset.}\label{Fig: TopWords}
\end{figure}

Figure \ref{Fig: TopWords} illustrates the agreement between the feature importance generated by our proposed PLEX model and those provided by LIME for emotion detection task. The bar chart displays the average overlap percentages between the top k important words identified by each method, calculated across all sentences in the dataset. Nearly 80\% of the time, the top word predicted by our proposed model and LIME is the same. However, when considering the top two words, the overlap reduces to 55\%, and for the top three words, it further drops to around 50\%.  Interestingly, beyond this initial dip, the overlap increases steadily as more words are included in the analysis. This trend suggests that while there might be slight discrepancies in the ranking of the most influential individual words, both methods demonstrate a converging consensus on the broader set of important features. This observation indicates that our model and LIME capture similar overall trends in feature importance, despite potential differences in their specific attribution mechanisms. 

\subsection{Comparing the polarity distribution}
\begin{figure*}
\centering
\frame{
\begin{subfigure}{0.3\textwidth}
  \includegraphics[width=\linewidth]{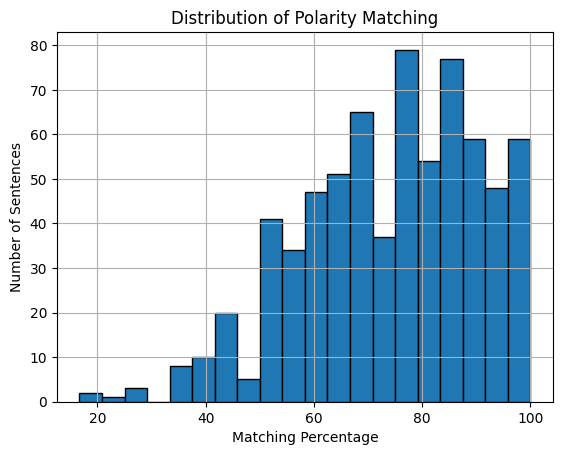}
  \caption{All words}
\end{subfigure}
\hfill
\begin{subfigure}{0.3\textwidth}
  \includegraphics[width=\linewidth]{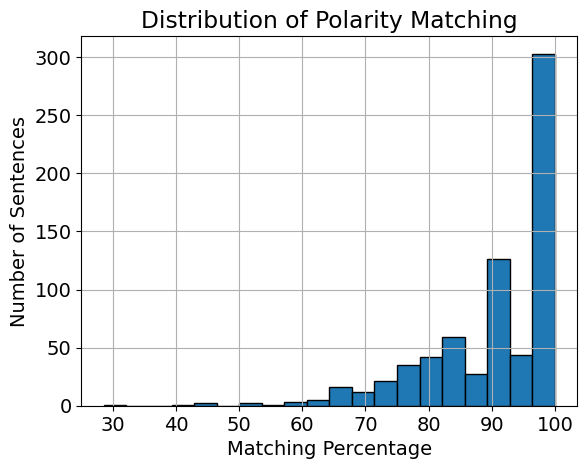}
  \caption{Words with Absolute score $>$ 0.01}
\end{subfigure}
\hfill
\begin{subfigure}{0.3\textwidth}
  \includegraphics[width=\linewidth]{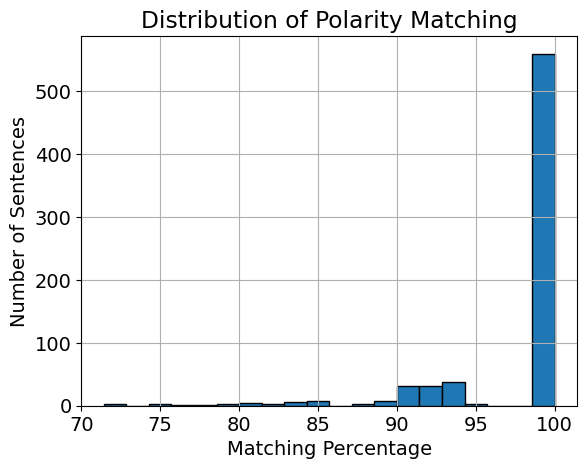}
  \caption{Words with Absolute score $>$ 0.05}
\end{subfigure}
}
\caption{Distribution of matching percentages between SHAP and PLEX feature importance scores}
\label{Fig: Polarity}
\end{figure*}

Figure \ref{Fig: Polarity} shows the agreement in feature importance scores generated by SHAP and the proposed PLEX  methods for depression classification task. By comparing the polarity (positive or negative sign) of the scores for each word across different sentences, we assessed the consistency between these two explanation techniques.  When we consider all the words, more than 75\% of the words in each sentence have the same polarity. However, when we consider the words with absolute importance score $0.01$ and $0.05$ or bigger, this agreement increased to $91\%$ and $98\%$. This suggests that both methods provide similar insights into the model's behavior, despite their different underlying approaches to explaining feature importance. We observed the same distribution for LIME across all use cases.

\subsection{Visualisation of all four methods}
\begin{figure*}\centering
\frame{\includegraphics[scale=0.5]{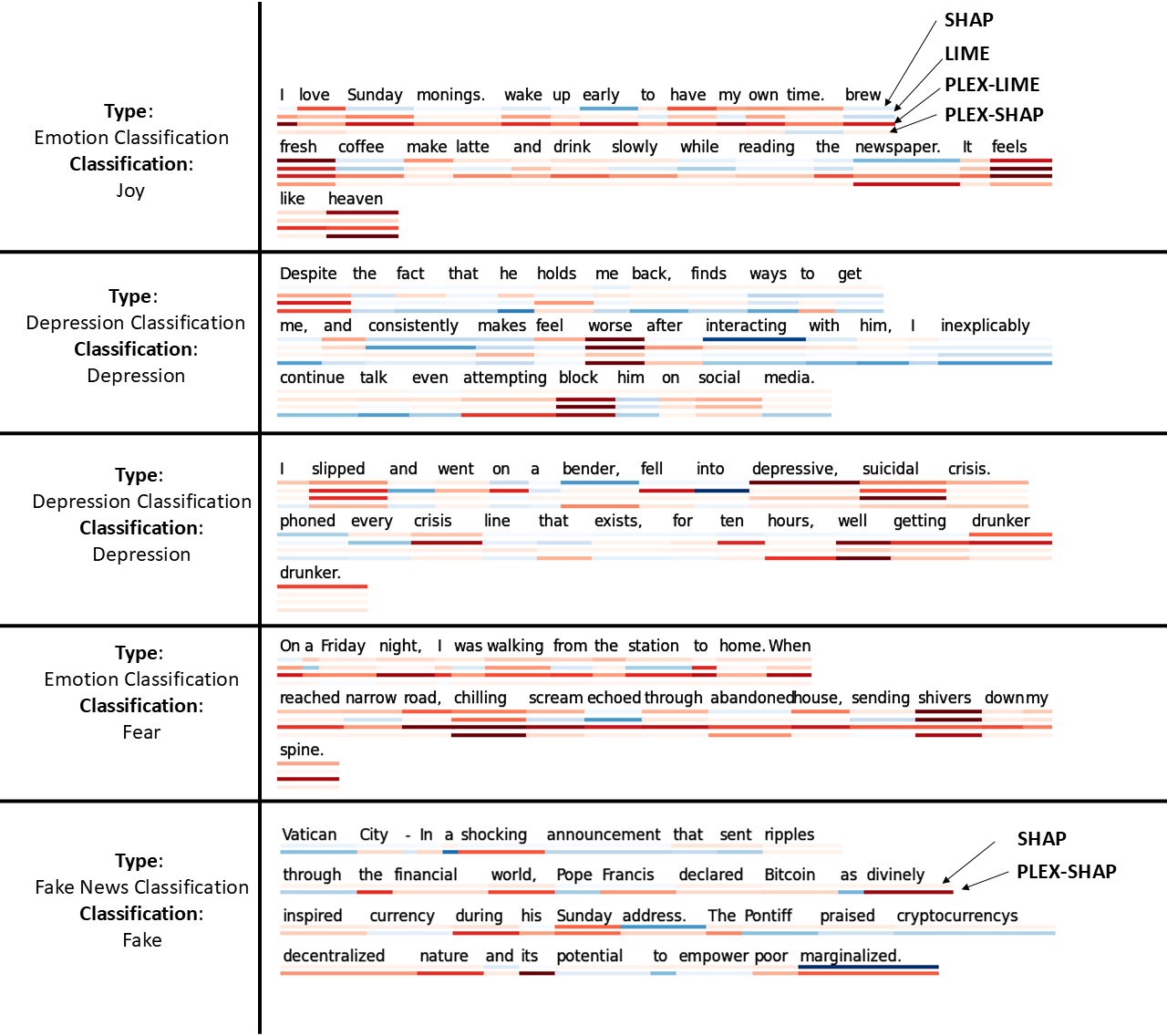}}
\caption{Comparison of word-level feature importance across four methods (SHAP, LIME, PLEX-LIME, and PLEX-SHAP. Positive contributions are highlighted in red, while negative contributions are shown in blue, with intensity proportional to importance magnitude.}\label{Fig: word_imp_hm}
\end{figure*}

Figure \ref{Fig: word_imp_hm} presents a comparison of word-level feature importance for five sentences, analyzed using emotion, depression, and fake news classifiers. Each word is annotated with horizontal lines to represent its importance as determined by four methods: SHAP, LIME,PLEX-LIME, and PLEX-SHAP. Red lines highlight positive contributions, and blue lines indicate negative contributions, with colour intensity reflecting the importance magnitude. The majority of the words show alignment across the four methods, demonstrating consensus. However, certain phrases, such as ``brew fresh coffee" and ``reading the newspaper," illustrate instances where the proposed methods outperform, capturing context-specific nuances more effectively than traditional approaches.

\subsection{Visualisong PLEX via Heatmap}
Figure \ref{Fig: ExamplHeatmap} showcases the word-level importance scores for an example sentence classified as indicating depression. Words with significance are highlighted in red and the word which not important are highlighted in blue. One key advantage of the proposed model is its ability to effectively highlight word importance even for significantly longer sentences, such as this example. In contrast, using traditional methods like LIME or SHAP would lead to exponentially increasing computational complexity as sentence length grows, making them less practical for such cases.

\subsection{Computational Complexity}
Table \ref{tab:time-comp-cost} presents a detailed comparison of computational efficiency and runtime across SHAP, LIME, and the proposed PLEX explanation methods for four tasks: T1/T4 (emotion/depression classification), T2 (fake news detection), and T3 (depression detection). Since T1 and T4 share the same underlying architecture, their results are grouped. Evaluations span three sentence length categories (small, medium, long) and, for SHAP and LIME, various perturbation sizes (256 to 8192) and batch sizes (16 to 512). PLEX, being perturbation-free, exhibits consistently lower computational cost.  As shown in the table, SHAP and LIME demonstrate a significant increase in both runtime and FLOPs (billions of floating-point operations) as the number of perturbations rises, particularly for longer sentences. While T1/T4 and T2 show manageable runtimes for shorter sentences, T3, with its more complex architecture (1024-dimensional embeddings and 24 layers), exhibits the highest computational cost, making SHAP and LIME computationally prohibitive for real-time applications in this context.

For small sentences, PLEX processes explanations in under 1 second (e.g., 0.52 seconds for T1/T4) compared to 5.67 seconds for SHAP and LIME at 256 perturbations. Even for long sentences, PLEX maintains high computational efficiency, with runtime not exceeding 3.01 seconds for the most complex model (T3). Similarly, PLEX significantly reduces FLOPs, scaling minimally with sentence length. For T3, PLEX achieves a computational cost of 11.21G FLOPs for long sentences, approximately three orders of magnitude lower than SHAP or LIME with 4096 perturbations (40.8T FLOPs).

The results in Table \ref{tab:time-comp-cost} reveal a stark contrast between PLEX and traditional SHAP/LIME explanation methods:
\begin{itemize}
    \item PLEX drastically reduces both runtime and computational cost by bypassing perturbation-based sampling, making it highly suitable for real-time applications and resource-constrained environments.
    \item  As sentence lengths increase, SHAP and LIME face exponential growth in computational requirements due to the expanding perturbation space. In contrast, PLEX scales gracefully across sentence lengths and models.
    \item PLEX demonstrates robust and consistent performance across all models and sentence lengths, offering a reliable explanation method for diverse text classification tasks.
\end{itemize}

The proposed PLEX method represents a significant leap forward in computational efficiency for explainable AI. By avoiding the limitations of perturbation-based sampling, PLEX provides a scalable and resource-efficient alternative to SHAP and LIME, making it particularly well-suited for modern NLP applications that demand both interpretability and practicality at scale.

\begin{table*}\centering
\caption{Comparison of PLEX and SHAP/LIME in terms of execution time (seconds) and FLOPs (GigaFLOPs) for explaining sentences of varying lengths using different pre-trained language models.}
\label{tab:time-comp-cost}
\begin{tabular}{|ccclcccccc|}
\hline
\multicolumn{4}{|l|}{\multirow{2}{*}{}}                                                                                                                   & \multicolumn{6}{c|}{Sentence Length}                                                                                                                                                                                                 \\ \cline{5-10} 
\multicolumn{4}{|l|}{}                                                                                                                                    & \multicolumn{2}{c|}{Small (avg. 10 words)}                                        & \multicolumn{2}{c|}{Medium (avg. 20 words)}                                       & \multicolumn{2}{c|}{Long (avg. 30 words)}                    \\ \hline
\multicolumn{1}{|c|}{\multirow{8}{*}{SHAP/LIME}} & \multicolumn{3}{c|}{\begin{tabular}[c]{@{}c@{}}Number of\\ Purturbation\end{tabular}}                  & \multicolumn{1}{c|}{256}                & \multicolumn{1}{c|}{512}                & \multicolumn{1}{c|}{1024}               & \multicolumn{1}{c|}{2048}               & \multicolumn{1}{c|}{4096}               & 8192               \\ \cline{2-10} 
\multicolumn{1}{|c|}{}                           & \multicolumn{3}{c|}{Batch Size}                                                                        & \multicolumn{1}{c|}{16}                 & \multicolumn{1}{c|}{32}                 & \multicolumn{1}{c|}{64}                 & \multicolumn{1}{c|}{128}                & \multicolumn{1}{c|}{256}                & 512                \\ \cline{2-10} 
\multicolumn{1}{|c|}{}                           & \multicolumn{1}{c|}{\multirow{3}{*}{Time (s)}}                  & \multicolumn{2}{c|}{T1/T4}           & \multicolumn{1}{c|}{5.67}               & \multicolumn{1}{c|}{6.34}               & \multicolumn{1}{c|}{7.13}               & \multicolumn{1}{c|}{7.39}               & \multicolumn{1}{c|}{40.77}              & 113                \\ \cline{3-10} 
\multicolumn{1}{|c|}{}                           & \multicolumn{1}{c|}{}                                           & \multicolumn{2}{c|}{T2}              & \multicolumn{1}{c|}{5.44}               & \multicolumn{1}{c|}{6.55}               & \multicolumn{1}{c|}{7.29}               & \multicolumn{1}{c|}{7.53}               & \multicolumn{1}{c|}{36.91}              & 117                \\ \cline{3-10} 
\multicolumn{1}{|c|}{}                           & \multicolumn{1}{c|}{}                                           & \multicolumn{2}{c|}{T3}              & \multicolumn{1}{c|}{11.43}              & \multicolumn{1}{c|}{12,32}              & \multicolumn{1}{c|}{12.52}              & \multicolumn{1}{c|}{13.44}              & \multicolumn{1}{c|}{33.45}              & 74.35              \\ \cline{2-10} 
\multicolumn{1}{|c|}{}                           & \multicolumn{1}{c|}{\multirow{3}{*}{Flops (G)}}                 & \multicolumn{2}{c|}{T1/T4}           & \multicolumn{1}{c|}{261}                & \multicolumn{1}{c|}{523}                & \multicolumn{1}{c|}{1917}               & \multicolumn{1}{c|}{3829}               & \multicolumn{1}{c|}{11147}              & 22294              \\ \cline{3-10} 
\multicolumn{1}{|c|}{}                           & \multicolumn{1}{c|}{}                                           & \multicolumn{2}{c|}{T2}              & \multicolumn{1}{c|}{261}                & \multicolumn{1}{c|}{523}                & \multicolumn{1}{c|}{1917}               & \multicolumn{1}{c|}{3829}               & \multicolumn{1}{c|}{11496}              & 23001              \\ \cline{3-10} 
\multicolumn{1}{|c|}{}                           & \multicolumn{1}{c|}{}                                           & \multicolumn{2}{c|}{T3}              & \multicolumn{1}{c|}{929}                & \multicolumn{1}{c|}{1857}               & \multicolumn{1}{c|}{6810}               & \multicolumn{1}{c|}{13606}              & \multicolumn{1}{c|}{40818}              & 81636              \\ \hline
\multicolumn{10}{|l|}{}                                                                                                                                                                                                                                                                                                                                                                          \\ \hline
\multicolumn{1}{|c|}{\multirow{8}{*}{PLEX}}      & \multicolumn{3}{c|}{\multirow{2}{*}{\begin{tabular}[c]{@{}c@{}}Number of\\ Purturbation\end{tabular}}} & \multicolumn{1}{c|}{\multirow{2}{*}{0}} & \multicolumn{1}{c|}{\multirow{2}{*}{0}} & \multicolumn{1}{c|}{\multirow{2}{*}{0}} & \multicolumn{1}{c|}{\multirow{2}{*}{0}} & \multicolumn{1}{c|}{\multirow{2}{*}{0}} & \multirow{2}{*}{0} \\
\multicolumn{1}{|c|}{}                           & \multicolumn{3}{c|}{}                                                                                  & \multicolumn{1}{c|}{}                   & \multicolumn{1}{c|}{}                   & \multicolumn{1}{c|}{}                   & \multicolumn{1}{c|}{}                   & \multicolumn{1}{c|}{}                   &                    \\ \cline{2-10} 
\multicolumn{1}{|c|}{}                           & \multicolumn{1}{c|}{\multirow{3}{*}{Time (s)}}                  & \multicolumn{2}{c|}{T1/T4}           & \multicolumn{1}{c|}{0.52}               & \multicolumn{1}{c|}{0.52}               & \multicolumn{1}{c|}{0.54}               & \multicolumn{1}{c|}{0.54}               & \multicolumn{1}{c|}{0.55}               & 0.55               \\ \cline{3-10} 
\multicolumn{1}{|c|}{}                           & \multicolumn{1}{c|}{}                                           & \multicolumn{2}{c|}{T2}              & \multicolumn{1}{c|}{1.25}               & \multicolumn{1}{c|}{1.25}               & \multicolumn{1}{c|}{1.32}               & \multicolumn{1}{c|}{1.32}               & \multicolumn{1}{c|}{2.01}               & 2.01               \\ \cline{3-10} 
\multicolumn{1}{|c|}{}                           & \multicolumn{1}{c|}{}                                           & \multicolumn{2}{c|}{T3}              & \multicolumn{1}{c|}{2.01}               & \multicolumn{1}{c|}{2.01}               & \multicolumn{1}{c|}{2.23}               & \multicolumn{1}{c|}{2.23}               & \multicolumn{1}{c|}{3.01}               & 3.01               \\ \cline{2-10} 
\multicolumn{1}{|c|}{}                           & \multicolumn{1}{c|}{\multirow{3}{*}{Flops (G)}}                 & \multicolumn{2}{c|}{T1/T4}           & \multicolumn{1}{c|}{1.02}               & \multicolumn{1}{c|}{1.02}               & \multicolumn{1}{c|}{1.87}               & \multicolumn{1}{c|}{1.87}               & \multicolumn{1}{c|}{2.72}               & 2.72               \\ \cline{3-10} 
\multicolumn{1}{|c|}{}                           & \multicolumn{1}{c|}{}                                           & \multicolumn{2}{c|}{T2}              & \multicolumn{1}{c|}{3.62}               & \multicolumn{1}{c|}{3.62}               & \multicolumn{1}{c|}{6.65}               & \multicolumn{1}{c|}{6.65}               & \multicolumn{1}{c|}{9.62}               & 9.62               \\ \cline{3-10} 
\multicolumn{1}{|c|}{}                           & \multicolumn{1}{c|}{}                                           & \multicolumn{2}{c|}{T3}              & \multicolumn{1}{c|}{4.68}               & \multicolumn{1}{c|}{4.68}               & \multicolumn{1}{c|}{7.07}               & \multicolumn{1}{c|}{7.07}               & \multicolumn{1}{c|}{11.21}              & 11.21              \\ \hline
\end{tabular}
\end{table*}

\section{Conclusions and Future Works}\label{Section: Conclusions}
In this paper, we introduced PLEX, a novel approach for generating efficient and interpretable local explanations for LLM-based text classification without relying on perturbation techniques or surrogate models. Our method leverages the rich contextual embeddings extracted from the LLM and employs a ``Siamese network" trained to align these embeddings with feature importance scores. This one-off training process eliminates the need for repeated inferences on perturbed inputs, significantly reducing the computational overhead associated with traditional XAI methods like LIME and SHAP. We demonstrated the effectiveness of PLEX using four different classification tasks, showing strong agreement with LIME and SHAP explanations while achieving a substantial reduction in computational cost. Our evaluation using a ``stress test" revealed that PLEX accurately captures the impact of removing important words on classification accuracy, further validating its faithfulness in identifying influential features. Notably, for certain tasks, PLEX even demonstrated superior performance compared to LIME and SHAP. The development of PLEX addresses a crucial need for efficient and scalable XAI methods for LLMs, particularly in applications where real-time explanations or limited computational resources are essential.

Future research directions include exploring the use of context-specific word embeddings, rather than the CLS token embedding, as input to the Siamese network. This approach could potentially enhance the network's ability to capture fine-grained nuances in sentiment expression. For instance, in emotion classification, utilizing the embeddings of words like ``happy" or ``sad" might provide more targeted information about the specific emotion being conveyed. Additionally, investigating the effectiveness of different similarity measures and network architectures within the PLEX framework could further enhance its performance and generalizability.

\ifCLASSOPTIONcaptionsoff
  \newpage
\fi

\balance

\end{document}